# Randomized ICA and LDA Dimensionality Reduction Methods for Hyperspectral Image Classification

Chippy Jayaprakash[1], Bharath Bhushan Damodaran[2], *Member, IEEE*, Sowmya V[1] and K P Soman[1]

*Abstract*—Dimensionality reduction is an important step in processing the hyperspectral images (HSI) to overcome the curse of dimensionality problem. Linear dimensionality reduction methods such as Independent component analysis (ICA) and Linear discriminant analysis (LDA) are commonly employed to reduce the dimensionality of HSI. These methods fail to capture non-linear dependency in the HSI data, as data lies in the nonlinear manifold. To handle this, nonlinear transformation techniques based on kernel methods were introduced for dimensionality reduction of HSI. However, the kernel methods involve cubic computational complexity while computing the kernel matrix, and thus its potential cannot be explored when the number of pixels (samples) are large. In literature a fewer number of pixels are randomly selected to partial to overcome this issue, however this sub-optimal strategy might neglect important information in the HSI. In this paper, we propose randomized solutions to the ICA and LDA dimensionality reduction methods using Random Fourier features, and we label them as RFFICA and RFFLDA. Our proposed method overcomes the scalability issue and to handle the non-linearities present in the data more efficiently. Experiments conducted with two real-world hyperspectral datasets demonstrates that our proposed randomized methods outperform the conventional kernel ICA and kernel LDA in terms overall, per-class accuracies and computational time.

## I. INTRODUCTION

Hyperspectral Images (HSIs) are captured by imaging spectroscopy that constitutes both spectral and spatial information [1]. HSI is made out of hundreds of bands with a high spectral resolution, ranging from visible to infrared region. Every pixel of a hyperspectral image relates to one of the materials present on the surface of the Earth [2]. For a single pixel, the reflectance values of all the spectral bands present in a hyperspectral image constitutes the spectral signature of the corresponding material. The concept of spectral signature enables to characterize the unique materials present on the Earth's surface and thus makes the HSI as an appropriate candidate for land cover classification. This has prompted the broad utilization of hyperspectral images as an imperative information for various applications such as natural observation, vegetation health monitoring, mineral investigation, target identification useful for defence, military, etc. [3], [4], [5], [6]. The processing of HSI is necessary to extract the information required for various applications [7], [8]. The challenges in processing the hyperspectral images are large data size and the presence of redundant information due to the strong correlation between the spectral bands. Therefore, the direct processing of HSI leads to the curse of high dimensional data processing. Hence, the pre-processing of HSI is necessary to benefit its potential in real-world applications . Dimensionality reduction (feature extraction) and feature selection is commonly employed to reduce the dimensionality of HSI data by mapping the original data to a lower dimensional space or by selecting subset of bands without loss of information [9].

In literature, several data transformation methods were proposed for dimensionality reduction. This includes Principal Component Analysis (PCA), Minimum Noise Fraction [10], Independent Component Analysis (ICA) [11], Linear Discriminant Analysis (LDA) [12], etc. These methods fall under the category of linear transformation methods [13]. Each of these methods extract features based on the existence of linear relationships in the data. For example, PCA deals with the data variance whereas, MNF works based on Signal to Noise Ratio. LDA is based on maximizing the ratio of between-class to within-class scatter in the data. In case of ICA, it is assumed that each band is a linear mixture of some concealed components and thus, a linear unmixing strategy is used to separate the independent components.

In any case, the regular assumption about HSI is that the data lies in a nonlinear complex manifold within the original dataset. All the existing linear transformation methods fail to capture the nonlinearity present in the data. Hence to extract the nonlinear relationships, the conventional methods were refined to kernel versions to form Kernel Principal Component Analysis (KPCA) [14], Kernel Minimum Noise Fraction (KMNF) [15], Kernel Independent Component Analysis (kernel-ICA) [16] and Generalized Discriminant Analysis (GDA) [17]. In this kernel strategy, a mapping capacity is utilized to delineate information to Kernel Hilbert space, which includes computation of a high dimensional piece grid. Since an expansive number of pixels constitute the HSI, figuring kernel matrix summons cubic computational complexity. Hence, only a few samples are selected to construct the kernel matrix [14], [15]. The results obtained through the kernel methods may be of less quality, as it involves only a small subset of samples. In order to overcome this limitation, Ali Rahimi and Ben Recht presented a Random Fourier Feature map to uncover the nonlinear relationship present in the data [18]. This technique finds an unequivocal low dimensional component to map kernel matrix. Recently, the estimate of KPCA utilizing Random Fourier Features (RFF) is presented in [19] and the randomized methods of PCA and MNF are proposed for the dimensionality reduction of hyperspectral data in [20]. But, as PCA depends on the second order statistics, it may fail to characterize many delicate materials that are caught by high-resolution HSI sensor. Also, the features extracted through PCA is not always independent and invariant under change, which may negate the assumptions made for many supervised classifications [21]. These drawbacks of PCA were overcome by ICA, which captures the linear relationships in the data with the help of higher order statistics. Later, even though kernel ICA was introduced to capture nonlinearity, it involved high computational time and storage due to the implicit lifting by kernel trick while processing the data. With the evolution of randomized feature maps, it was possible to explicitly map the higher dimensional data to a low dimensional Euclidean inner product space by saving storage space and computational time [18], [22].

[1]Chippy Jayaprakash, Sowmya V and K P Soman are with Center for Computational Engineering and Networking (CEN), Amrita School of Engineering, Coimbatore, Amrita Vishwa Vidyapeetham, India. chippyjprakash@gmail.com, v_sowmya@cb.amrita.edu, kp_soman@amrita.edu

[2]Bharath Bhushan Damodaran is with Univ. Bretagne-Sud, UMR 6074, IRISA, F-56000 Vannes, France. bharath-bhushan.damodaran@irisa.fr

In this paper, we contribute RFF based non-linear analysis i.e., randomized ICA (RFFICA) and randomized LDA (RFFLDA) for the dimensionality reduction of hyperspectral images, which maps the data to a higher dimension using RFF maps. The proposed methods overcome the cost of computational time of kernel methods by finding a low dimensional RFF based feature map to approximate the kernel matrix. The performance of the proposed methods over conventional and kernel methods are evaluated based on classification accuracy and computational time, which is experimented on two standard hyperspectral datasets.

The rest of this paper is organized in the following manner: section II describes background theory required for the proposed methods and section III presents the proposed method. Section IV describes the experimental setup, data-sets and the baseline approaches used in the present work. Section V reports the experimental results and discussion and the conclusion derived from the present work is given in Section VI.

## II. Background

In this section, we briefly describe about independent component analysis, linear discriminant analysis and random Fourier features methods which are necessary to present our proposed method.

### A. Independent Component Analysis (ICA)

Let us consider a mixture of random variables $\mathbf{x}_1, \mathbf{x}_2, ... \mathbf{x}_N$, where each $\mathbf{x}_i \in \mathbb{R}^d$. These random variables are defined as a linear combination of another random variables $\mathbf{p}_1, \mathbf{p}_2, \ldots, \mathbf{p}_N$, where each $\mathbf{p}_i \in \mathbb{R}^n$. Then, the mixing model can be mathematically written as,

$$\boldsymbol{X} = \boldsymbol{AP} \quad (1)$$

where $\boldsymbol{X} = [\mathbf{x}_1, \mathbf{x}_2, \ldots, \mathbf{x}_N]$ is the observed vector, $\boldsymbol{P} = [\mathbf{p}_1, \mathbf{p}_2, \ldots, \mathbf{p}_N]$ is the unknown source, $\boldsymbol{A}$ is the mixing matrix, $n$ gives the number of unknown sources and $d$ is the number of observations made. In order to find the independent components, we need to solve for the unmixing matrix $\boldsymbol{W}$ (inverse of $\boldsymbol{A}$). The independent components are obtained using below given equation 2.

$$ICA(\boldsymbol{X}) = \boldsymbol{P} = \boldsymbol{A}^{-1}\boldsymbol{X} = \boldsymbol{WX} \quad (2)$$

If we consider $\boldsymbol{X} \in \mathbb{R}^{d \times N}$ as the hyperspectral image we get,

$$\boldsymbol{P}_{n \times N} = \boldsymbol{W}_{n \times d} \boldsymbol{X}_{d \times N} \quad (3)$$

where, $N$ is the number of pixels in each band, $d$ represents the number of spectral bands and $n$ gives the number of sources or materials present in the image.

The estimation of the ICA model is conceivable, only if the accompanying presumptions and limitations are fulfilled: (i) statistically independent sources; (ii) independent components should possess non Gaussian distribution; (iii) $\boldsymbol{A}$ should be a square and full rank matrix.

### B. Linear Discriminant Analysis (LDA)

LDA is a supervised dimensionality reduction method, which requires labeled samples to estimate the transformation matrix. Let's consider a set of given labeled samples $\{\mathbf{x}_i, l_i\}_{i=1}^N$, $\mathbf{x}_i \in \mathbb{R}^d$, $l_i \in \Omega = \{\omega_1, \omega_2, \ldots, \omega_C\}$. Each $\mathbf{x_i} \in \mathbb{R}^d$ is the $d$-dimensional element vector of the $i^{th}$ pixel with class label $l_i \in \Omega$. Here, $d$ signifies the number of spectral bands and $\Omega$ characterizes the universe of all conceivable labeled classes in the image. Let $N_j$ be the number of samples belonging to class $\omega_j$ and $\mu_\mathbf{j}$ be the mean vector of each class.

The standard LDA classifier enables us to locate a direct transformation matrix G that diminishes a $d$ dimensional feature vector $\mathbf{x}$ to a $q$ dimensional vector, $\mathbf{f} = \boldsymbol{G}^T \mathbf{x} \in \mathbb{R}^q$ such that $q < d$. $\boldsymbol{G} = [\mathbf{g}_1, \mathbf{g}_2, \ldots, \mathbf{g}_C]^T$ is the projection matrix in which, each $\mathbf{g}_i \in \mathbb{R}^d$. The Fisher criterion depends on maximizing the distance between mean of classes and simultaneously limiting their intraclass fluctuations based on the equation 4:

$$J(\boldsymbol{G}) = \frac{\boldsymbol{G}^T \boldsymbol{S}_b \boldsymbol{G}}{\boldsymbol{G}^T \boldsymbol{S}_w \boldsymbol{G}} \quad (4)$$

Based on the decision function $\mathbf{y} = \boldsymbol{W}^T \mathbf{x}$ The above equation can be re-written as :

$$\boldsymbol{G}^* = \arg\max \frac{|\boldsymbol{G}^T \boldsymbol{S}_b \boldsymbol{G}|}{|\boldsymbol{G}^T \boldsymbol{S}_w \boldsymbol{G}|} \quad (5)$$

$\boldsymbol{S_w}$ is the within-class scatter which is given by $\boldsymbol{S_w} = \sum_{j=1}^{C} \boldsymbol{S}_j$ where, $\boldsymbol{S}_j = \sum_{\mathbf{x_i} \in \omega_j}(\mathbf{x_i} - \mu_\mathbf{j})(\mathbf{x_i} - \mu_\mathbf{j})^T$ in which $\mu_\mathbf{j} = \frac{1}{N_j} \sum_{\mathbf{x_i} \in \omega_j} \mathbf{x_i}$. The between-class scatter is given by $\boldsymbol{S_b} = \sum_{j=1}^{C} N_j(\mu_\mathbf{j} - \mu)(\mu_\mathbf{j} - \mu)^T$ where, $\mu = \frac{1}{N} \sum_{i=1}^{N} \mathbf{x_i}$.

In this $C$ is the total number of classes in the data. We get the optimal projection matrix $\boldsymbol{G}^*$, whose columns are the eigenvectors corresponding to the largest eigenvalues of the eigenvalue problem:

$$(\boldsymbol{S_b} - \lambda_j \boldsymbol{S_w})\mathbf{g_j} = 0 \quad (6)$$

In the next subsection, we briefly describes the kernel approximation using Random Fourier Features.

### C. Kernel Approximation using Random Fourier Features

Kernel methods are used as an efficient strategy to find the non linear relationships in the data. It projects the data to a higher dimensional space for capturing the features i.e., for any $\mathbf{x}, \mathbf{y} \in \mathbb{R}^d$:

$$\mathbf{K}(\mathbf{x}, \mathbf{y}) = <\phi(\mathbf{x}), \phi(\mathbf{y})> \quad (7)$$

where $\phi$ is the mapping function to the RKHS space. Kernel methods are expensive in computation, when dealing with HSI (or large scale data) due to its huge number of pixels. To avoid the complexity, kernel approximation seeks to find a low dimensional feature map that approximates the kernel matrix [18].

$$\mathbf{K}(\mathbf{x}, \mathbf{y}) = <\phi(\mathbf{x}), \phi(\mathbf{y})> \simeq \mathbf{z}(\mathbf{x})^\mathbf{T} \mathbf{z}(\mathbf{y}) \quad (8)$$

For better understanding let us approximate the RBF kernel. The RBF kernel is defined as

$$\mathbf{K}(\mathbf{x}, \mathbf{y}) = e^{-\gamma\|\mathbf{x}-\mathbf{y}\|_2^2} = e^{-\gamma(\mathbf{x}-\mathbf{y})^\mathbf{T}(\mathbf{x}-\mathbf{y})} = e^{-\gamma(\mathbf{z}^\mathbf{T}\mathbf{z})} = p(\mathbf{z}) \quad (9)$$

By Bocheners theorem, Fourier transform of a shift invariant kernel is a positive measure:

$$\mathbf{K}(\mathbf{x} - \mathbf{y}) = \int P(\mathbf{r}) e^{\mathbf{r}^\mathbf{T}(\mathbf{x}-\mathbf{y})} d(\mathbf{r})$$

$$= E(e^{-j\mathbf{r}^\mathbf{T}(\mathbf{x}-\mathbf{y})})$$

$$\simeq \frac{2}{D} \sum_{i=1}^{D} \cos(\mathbf{r}_i^T \mathbf{x} + b_i)(\cos(\mathbf{r}_i^T \mathbf{y} + b_i) = \mathbf{z}(\mathbf{x})^T \mathbf{z}(\mathbf{y})$$

where

$$\mathbf{z}(\mathbf{x}) = \sqrt{\frac{2}{D}}[\cos(\mathbf{r}_1^T \mathbf{x} + b_1), ....\cos(\mathbf{r}_D^T \mathbf{x} + b_D)] \quad (10)$$

in which, $D$ is the number of Random Fourier features. The RBF kernel is approximated by sampling $\mathbf{r}$ from N(0, $\sigma^{-2}$) and $b$ is drawn from uniform distribution [18], [20].

In this paper, we propose scalable nonlinear component analysis methods namely RFFICA and RFFLDA for the dimensionality reduction of hyperspectral images by using the feature map mentioned in equation 10.

## III. PROPOSED METHODS

Consider $\boldsymbol{X} \in \mathbb{R}^{d \times N}$ as the hyperspectral image and let $\boldsymbol{R} \in \mathbb{R}^{d \times D}$ be the RFF coefficients. Then, the randomized methods of ICA and LDA can be performed on the hyperspectral data as follows:

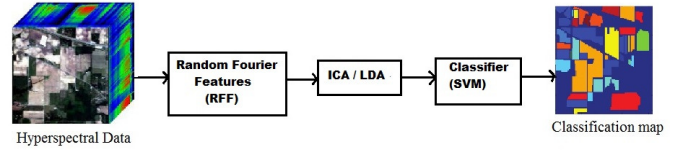

Fig. 1: Block diagram of the proposed work

### A. Randomized ICA (RFFICA)

1) Compute $\boldsymbol{I} = \mathbf{z}(X)$ using random feature map given in equation 10, which maps the input HSI data of $\mathbb{R}^{d \times N}$ to $\mathbb{R}^{D \times N}$ space.
2) Perform ICA using equation (2) on the result obtained from step 1.

This results in a matrix $\boldsymbol{P}$ of dimension $n \times N$. Here, $d$ is the number of spectral bands, $N$ is the number of pixels in each band of the data, $n$ is the number of sources in the data and $D$ is the number of random Fourier features. The obtained randomized ICA is an approximation to kernel-ICA.

$$RFFICA(X) = ICA(I) \simeq KICA(X) \quad (11)$$

### B. Randomized LDA (RFFLDA)

Similarly, the randomized method of LDA can be computed as given below :

1) Compute $\boldsymbol{I} = \mathbf{z}(\boldsymbol{X})$ using random feature map given in equation 10, that maps the HSI data to $\mathbb{R}^{D \times N}$ space.
2) Perform LDA on $\boldsymbol{I}$ i.e,

$$LDA(\boldsymbol{I}) = \boldsymbol{Y} = \boldsymbol{G}^T \boldsymbol{I} \quad (12)$$

This results in $q \times N$ dimension matrix $\boldsymbol{Y}$. Here, $\boldsymbol{G}$ is the matrix of dimension $D \times q$ in which, $q$ is the number of components required.
i.e.,

$$RFFLDA(\boldsymbol{X}) = LDA(\boldsymbol{I}) \simeq GDA(\boldsymbol{X}) \quad (13)$$

When we perform the kernel versions of ICA and LDA on HSI, it involves the computation of kernel matrix of dimension $N \times N$. This computation of kernel matrix is too expensive as the number of pixels in HSI data is very large. Whereas, when the randomized methods are performed, it involves computation of a matrix of dimension $D \times N$ (where $D << N$) which is less expensive with respect to kernel method.

Thus, the randomized methods RFFICA and RFFLDA can be viewed as the low rank approximation of the kernel methods of ICA and LDA respectively.

## IV. EXPERIMENTAL SETUP

### A. Dataset

In order to evaluate the effectiveness of our proposed method, experiments are conducted with two real world hyperspectral datasets from two different settings: agriculture and urban land cover.

- Salinas Scene: This scene was gathered by the 224-band AVIRIS sensor over Salinas Valley, California. The territory secured involves 512 lines by 217 samples, with spatial resolution of 3.7 meter/pixels. The 20 water absorption bands are disposed in the range: [108-112], [154-167], 224. Therefore total number of bands after removal is 204. It incorporates vegetables, exposed soils, and vineyard fields. Salinas groundtruth contains 16 classes. The classes of Salinas scene and their corresponding number of samples is listed in table I.
- Pavia University: The hyperspectral data considered here was collected over the University of Pavia, Italy by the ROSIS airborne hyperspectral sensor in the framework of the HySens project managed by DLR (German national aerospace agency). The ROSIS sensor collects images in 115 spectral bands in the spectral range from 0.43 to 0.86 $\mu$m with a spatial resolution of 1.3 m/pixel. After the removal of noisy bands, 103 bands were selected for experiments. This data contains $610 \times 340$ pixels with nine classes of interest. The groundtruth classes for Pavia University with their corresponding number of samples are given in table II.

TABLE I: Groundtruth classes for Salinas scene and their respective number of samples

| # | Class | Samples |
|---|---|---|
| 1 | Brocoli_green_weeds_1 | 2009 |
| 2 | Brocoli_green_weeds_2 | 3726 |
| 3 | Fallow | 1976 |
| 4 | Fallow_rough_plow | 1394 |
| 5 | Fallow_smooth | 2678 |
| 6 | Stubble | 3959 |
| 7 | Celery | 3579 |
| 8 | Grapes_untrained | 11271 |
| 9 | Soil_vinyard_develop | 6203 |
| 10 | Corn_senesced_green_weeds | 3278 |
| 11 | Lettuce_romaine_4wk | 1068 |
| 12 | Lettuce_romaine_5wk | 1927 |
| 13 | Lettuce_romaine_6wk | 916 |
| 14 | Lettuce_romaine_7wk | 1070 |
| 15 | Vinyard_untrained | 7268 |
| 16 | Vinyard_vertical_trellis | 1807 |

TABLE II: Groundtruth classes for the Pavia University scenes and their respective number of samples

| # | Class | Samples |
|---|---|---|
| 1 | Asphalt | 6631 |
| 2 | Meadows | 18649 |
| 3 | Gravel | 2099 |
| 4 | Trees | 3064 |
| 5 | Painted metal sheets | 1345 |
| 6 | Bare Soil | 5029 |
| 7 | Bitumen | 1330 |
| 8 | Self-Blocking Bricks | 3682 |
| 9 | Shadows | 947 |

## B. Competitors

Our proposed randomized method is compared with state-of-the-art methods in the literature. For the randomized ICA (RFFICA) method, we compare with linear ICA, and kernel ICA. Similarly for the randomized LDA (RFFLDA) method, we compare with linear LDA and generalized discriminant analysis (GDA). For more details about kernel ICA and GDA, please refer to [16][17]. As the kernel matrix cannot be computed for all the pixels in the hyperspectral image for the kernel ICA and GDA, we adopted the sampling strategy which is followed in the literature [14][15]. We randomly sampled 2000 samples to compute the kernel matrix for the kernel ICA and GDA. For the RFFICA and RFFLDA the random Fourier feature coefficients are sampled from $N(0, \sigma^{-2})$, where $\sigma^2$ is estimated based on mean of the pair-wise distance between the samples as mentioned [20]. The same bandwidth parameter is used for the kernel ICA and GDA. FastICA [23] method is used to compute independent components for the linear ICA and RFFICA. For the rest of methods were implemented in Matlab.

## C. Assessment of proposed method

The extracted features of our proposed method and the state-of-the-art method are assessed in terms of classification performance (overall accuracy), visual inspection and computational time. For the classification experiments, the support vector machines (SVM) with RBF kernel is utilized as a classifier. The cost function $C = 2^\alpha$, $\alpha = \{-5, -4, \ldots, 15\}$, and the bandwidth parameter of the RBF kernel $\gamma = 2^\beta, \beta = \{-15, -13, \ldots, 4, 5\}$ of the SVM classifier are automatically tuned using grid search and five fold cross validation approach. For the classification experiments, we randomly choose 100 samples per class as training samples and whatever remains are used as testing samples. In order to avoid the bias, the experiments are repeated five times, and the average overall accuracy and per-class accuracies are reported. The entire procedure included in the experiment is portrayed in Fig. 1.

## V. RESULTS AND DISCUSSION

In this section, we present the experimental results of our proposed methods and existing methods. We first begin by discussing about the classification performance, and then we analyze about the computational complexity interms of computational time, and finally we inspect the quality of extracted components

### A. Analysis of Classification Results

Firstly, we analyze the results of our proposed randomized ICA (RFFICA) method for the two datasets, and later we discuss with randomized LDA method.

*1) RFFICA:* Table III reports the classification accuracies obtained with our proposed RFFICA method, linear ICA, and kernelized ICA for the Salianas Scence and Pavia University datasets. For these methods, the classification performance are compared with each other upto 35 number of extracted components. When the Salinas scene dataset is considered, as expected non-linear versions of ICA performed better than linear ICA. Our proposed RFFICA method consistently outperforms the linear ICA and kernel ICA methods and the performance difference is higher in lower number of components compared to linear ICA method. Furthermore, the per-class accuracies computed with 25 number of components of Salinas scene with all the methods (see Table IV), shows that our method performs better in the difficult classes, for instance class no. 15 in table IV.

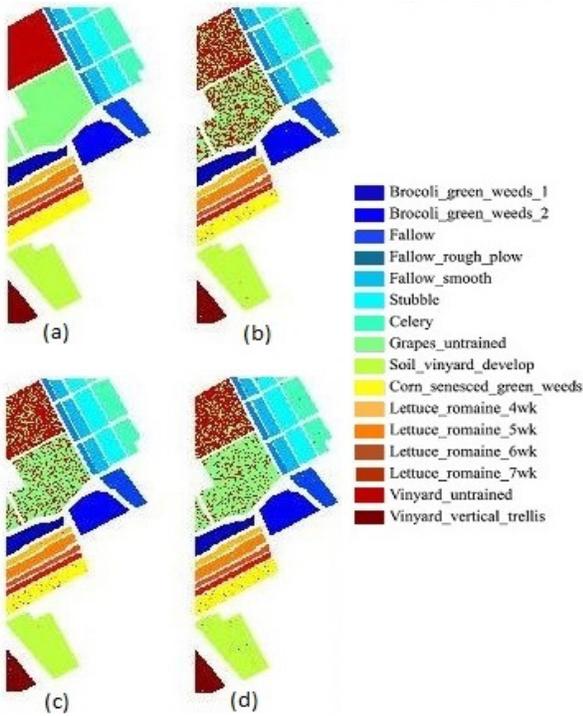

Fig. 2: (a) Ground truth of Salinas Scene. Classification map of Salinas Scene for 25 components using: (b) Linear ICA (c) Kernel ICA (d) RFFICA

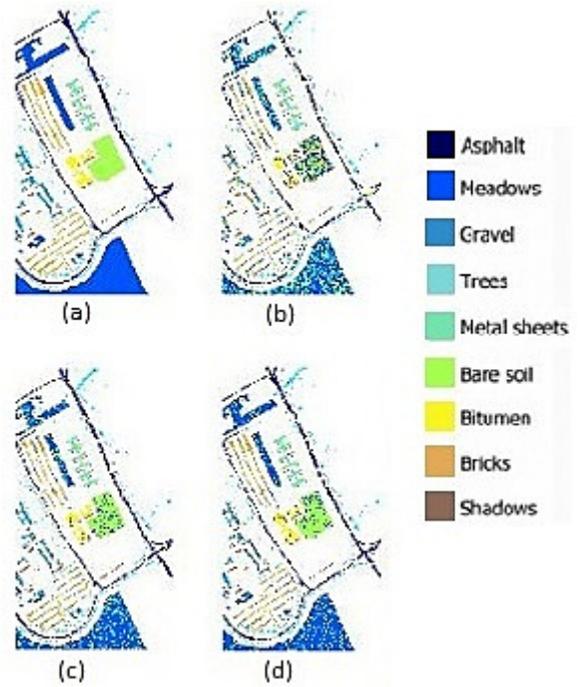

Fig. 3: (a) Ground truth of Pavia University. Classification map of Pavia University for 25 components using: (b) Linear ICA (c) Kernel ICA (d) RFFICA

Likewise, when Pavia University dataset is considered, our proposed RFFICA consistently outperforms the existing methods with larger margin over different number of extracted components. For example, with 10 number of extracted components, there is a difference about 5% and 30% approximately with respect to kernel ICA and linear ICA method. The classwise accuracies of all the methods with 25 components is listed in table V. It can be observed that, the randomized features have contributed to better classification of difficult classes, and it improved over 10% compared to the conventional kernel ICA.

The extracted number of components for all the methods is varied in the range of $3-35$. While it is below 3, only a few features would be extracted that results in lower classification accuracy. On the otherhand if it goes beyond 35, a presence of deviation in the accuracy rise occurs and also the kernel ICA and RFFICA method exhibits in similar performance as shown in table III. Finally, the classification maps are shown in Fig. 2 and 3, and visual inspection reveals that our proposed method offers better quality classification maps.

*2) RFFLDA:* Next, we discuss the performance of our proposed RFFLDA method. The LDA is a supervised dimensionality reduction method, unlike the ICA dimensionality reduction method. It is noted that for LDA method, number of extracted features can be only computed less than or equal to number of classes. The comparison of RFFLDA to other methods for both the datasets with different number of extracted components are provided in table VI and VIII. For the Salinas scene dataset, RFFLDA is slightly better than the LDA and GDA methods. In the lower order components (for e.g less than $4$), our method is better by above $7\%$ with respect to LDA method. Further the per-class accuracies with 6 number of components with different methods are given in table VII. It can be studied from table VII that the randomized features have contributed for better classification of each class of the Salinas Scene. The visual analysis of classification map for 6 components of Salinas is depicted in Fig. 4.

When the Pavia University dataset is considered, as observed with RFFICA, RFFLDA also outperforms the existing methods with large margin, especially when the number of components is considered above 6. Whereas when the number of components less than $4$ are analysed, we can see that the performance of the RFFLDA and GDA method is comparable. In the higher component, the performance of GDA method is similar to its linear counterpart, where as our method exhibit about $6\%$ improvement. This reveals that the randomized methods can be used for nonlinear component analysis to avoid the expensive computation and time as that of kernel methods. The visual analysis of classification for 6 components of PaviaU using the three methods for LDA can be seen from Fig. 5. It can be monitored from table IX that for PaviaU, the least exactness acquired in classification is for the class Meadows with 68.61% using GDA. But, for the same class, the accuracy obtained using randomized features is 73.94% that highlights the improvement in classification with an increase of almost 5%, which can be witnessed from table IX.

TABLE III: Comparison of Classification Accuracy (%) obtained using features of RFFICA against ICA and Kernel-ICA for two hyperspectral datasets. The reported accuracies are averaged over five runs and best accuracies are represented in **bold**

| Dataset | Method | Number of components | | | | | | | |
|---|---|---|---|---|---|---|---|---|---|
| | | 3 | 5 | 10 | 15 | 20 | 25 | 30 | 35 |
| Salinas Scene | ICA | 56.35 | 71.44 | 82.56 | 85.51 | 86.16 | 86.30 | 86.92 | 87.46 |
| | Kernel-ICA | **82.21** | 86.20 | **88.49** | 88.92 | 89.25 | 89.27 | 89.91 | 90.18 |
| | RFFICA | 79.67 | **86.30** | 88.38 | **88.98** | **89.69** | **90.00** | **90.07** | **90.29** |
| Pavia University | ICA | 23.32 | 26.90 | 33.42 | 38.99 | 48.24 | 51.88 | 54.74 | 57.24 |
| | Kernel-ICA | 57.04 | 67.09 | 74.37 | 78.65 | 79.79 | 81.44 | 81.67 | 82.72 |
| | RFFICA | **71.65** | **73.53** | **79.23** | **84.35** | **86.74** | **87.55** | **87.88** | **88.18** |

TABLE IV: Comparison of classwise accuracy (%) obtained for 25 components of Salinas scene dataset using ICA, Kernel-ICA and RFFICA. Best accuracies are reported in **bold**

| # | Class | Accuracy (%) | | |
|---|---|---|---|---|
| | | ICA | Kernel-ICA | RFFICA |
| 1 | Brocoli_green_weeds_1 | 97.43 | 99.06 | **99.27** |
| 2 | Brocoli_green_weeds_2 | 88.67 | 90.42 | **91.92** |
| 3 | Fallow | 82.54 | 84.63 | **85.04** |
| 4 | Fallow_rough_plow | 98.76 | **99.85** | 99.69 |
| 5 | Fallow_smooth | 82.16 | 84.29 | **85.17** |
| 6 | Stubble | 99.82 | 99.46 | **99.95** |
| 7 | Celery | 98.59 | 99.24 | **99.66** |
| 8 | Grapes_untrained | 56.14 | 60.60 | **65.01** |
| 9 | Soil_vinyard_develop | 88.15 | 90.02 | **92.55** |
| 10 | Corn_senesced_green_weeds | 78.20 | 80.67 | **82.80** |
| 11 | Lettuce_romaine_4wk | 93.70 | **98.04** | 93.70 |
| 12 | Lettuce_romaine_5wk | 81.27 | 83.58 | **84.28** |
| 13 | Lettuce_romaine_6wk | 97.06 | 98.53 | **99.51** |
| 14 | Lettuce_romaine_7wk | 96.60 | 93.87 | **94.20** |
| 15 | Vinyard_untrained | 50.64 | 67.01 | **71.82** |
| 16 | Vinyard_vertical_trellis | 97.42 | 98.71 | **98.81** |
| | **Average Accuracy** | 86.69 | 89.25 | **90.20** |

TABLE V: Comparison of classwise accuracy (%) obtained for 25 components of Pavia University dataset using ICA, Kernel-ICA and RFFICA. Best accuracies are reported in **bold**

| # | Class | Accuracy (%) | | |
|---|---|---|---|---|
| | | ICA | Kernel-ICA | RFFICA |
| 1 | Asphalt | 50.70 | 69.97 | **78.84** |
| 2 | Meadows | 39.76 | 74.55 | **90.09** |
| 3 | Gravel | 55.48 | 67.93 | **83.29** |
| 4 | Trees | 84.78 | 92.11 | **93.45** |
| 5 | Painted metal sheets | 99.40 | 99.20 | **99.60** |
| 6 | Bare Soil | 36.42 | 84.50 | **85.01** |
| 7 | Bitumen | 60.24 | 86.34 | **91.95** |
| 8 | Self-Blocking Bricks | 50.17 | 73.09 | **83.39** |
| 9 | Shadows | 96.81 | 99.88 | **99.91** |
| | **Average Accuracy** | 63.75 | 83.05 | **89.50** |

It uncovers that the features extracted through randomized

TABLE VI: Comparison of Classification Accuracy (%) obtained using features of RFFLDA against LDA and GDA for classification of Salinas Scene

| Dataset: Salinas Scene | | | | | | | |
|---|---|---|---|---|---|---|---|
| Method | Number of Components | | | | | | |
| | 2 | 4 | 6 | 8 | 10 | 12 | 15 |
| LDA | 63.04 | 77.06 | 85.20 | 87.23 | 88.55 | 88.87 | 88.86 |
| GDA | 74.82 | 84.84 | 86.31 | 87.72 | 87.52 | 88.99 | 89.73 |
| RFFLDA | **76.40** | **86.00** | **86.87** | **87.99** | **88.87** | **90.58** | **90.96** |

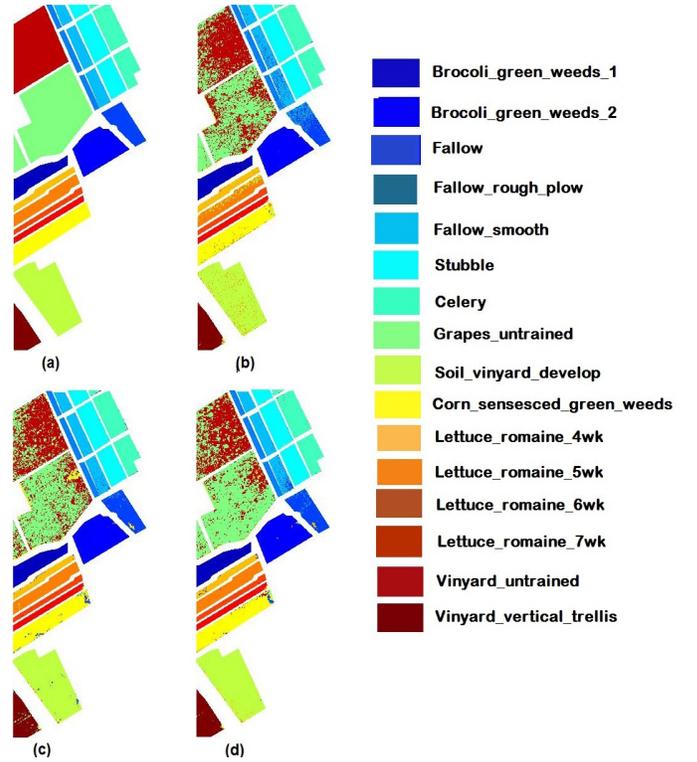

Fig. 4: (a) Ground truth of Salinas Scene. Classification map of Salinas scene for 6 components using:(b) LDA, (c) GDA, and (d) RFFLDA

strategy are more important for grouping of classes on the data, when contrasted with the features extricated through other methods. It also tells that, the performance of the method depends on the data being used. As in case of PaviaU for

TABLE VII: Classwise accuracy for 6 components of Salinas Scene dataset using LDA, GDA and RFFLDA

| # | Class | Accuracy (%) | | |
|---|---|---|---|---|
| | | LDA | GDA | RFFLDA |
| 1 | Brocoli_green_weeds_1 | **99.48** | 97.64 | 99.21 |
| 2 | Brocoli_green_weeds_2 | **99.75** | 98.84 | 99.17 |
| 3 | Fallow | 70.15 | 93.12 | **94.78** |
| 4 | Fallow_rough_plow | 87.02 | 99.57 | **99.88** |
| 5 | Fallow_smooth | 80.72 | 91.27 | **96.66** |
| 6 | Stubble | 99.52 | **99.61** | 99.43 |
| 7 | Celery | 99.04 | 99.11 | **99.40** |
| 8 | Grapes_untrained | 68.27 | 81.98 | **91.36** |
| 9 | Soil_vinyard_develop | 94.64 | 96.51 | **98.51** |
| 10 | Corn_senesced_green_weeds | 88.07 | 90.18 | **91.19** |
| 11 | Lettuce_romaine_4wk | 92.77 | 93.29 | **95.66** |
| 12 | Lettuce_romaine_5wk | 84.67 | 99.23 | **99.56** |
| 13 | Lettuce_romaine_6wk | **99.14** | 96.81 | 97.55 |
| 14 | Lettuce_romaine_7wk | 95.26 | 95.57 | **96.91** |
| 15 | Vinyard_untrained | 61.56 | 65.75 | **68.99** |
| 16 | Vinyard_vertical_trellis | 91.25 | 93.65 | **97.95** |
| | **Average Accuracy** | 88.20 | 93.25 | **95.39** |

TABLE VIII: Comparison of Classification Accuracy (%) obtained using features of RFFLDA against LDA and GDA for classification of Pavia University

| Dataset:Pavia University | | | | | | |
|---|---|---|---|---|---|---|
| Method | Number of Components | | | | | |
| | 2 | 3 | 4 | 6 | 8 | 9 |
| LDA | 47.40 | 65.33 | 70.64 | 73.16 | 75.72 | 78.16 |
| GDA | **57.55** | **70.50** | **72.98** | 77.05 | 78.66 | 78.00 |
| RFFLDA | 56.06 | 61.65 | 72.40 | **83.06** | **83.13** | **84.62** |

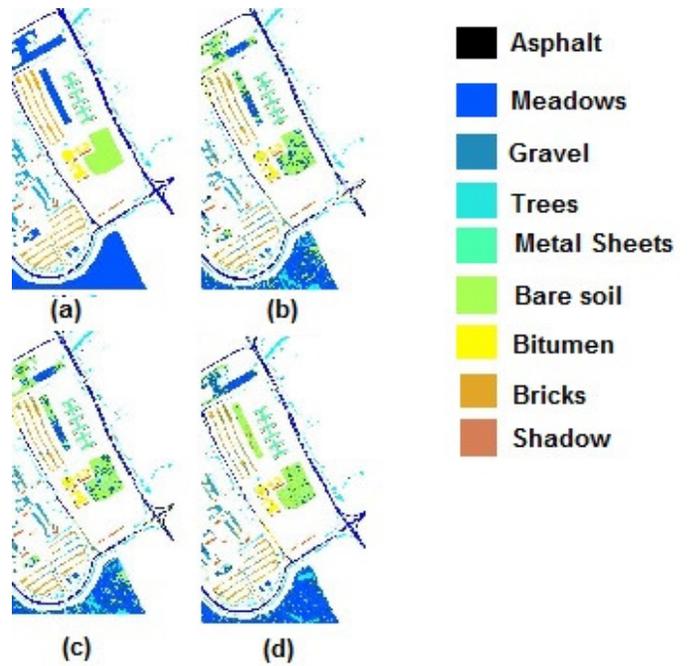

Fig. 5: (a) Ground truth of Pavia University. Classification map of Pavia University for 6 components using: (b) LDA, (c) GDA, and (d) RFFLDA

8 components, from table VIII, we can see that there is a contrast in the exactness acquired by randomized strategy to direct and kernel technique with an estimated distinction of 13% and 5% individually. While for Salinas scene of 8 components, there is just a practically identical distinction between the accuracies for RFFLDA, GDA and LDA. The features have added to the better classification of the classes, which shows the substance of significant data introduced in the features being extricated that describes a class. In specific, when upto 10 components are considered the improvement in the classification result obtained through randomized method is high for the two datasets that have been used for RFFICA and likewise for 8 components through RFFLDA, the performance is either higher than the other methods or comparable to kernel method.

### B. Visual Analysis of Extracted Components

We analyzed the effectiveness of proposed method through visual analysis of the extracted components. For this, Fig. 6 displays the first five extracted components of LDA, GDA and our proposed RFFLDA method. It is difficult to exactly quantify the quality of extracted components, however from the visual analysis it is evident RFFLDA has better quality components compared to existing methods. Furthermore, the Fig. 6 reveals that the different methods extracts different types of components.

### C. Computational time

In order to further demonstrate the advantages of our proposed method, here we compare the computational time complexity of all the methods, and these values are mentioned in table X. We restrict our analysis with ICA method using Salians Scene dataset. From the table X, we can infer that the linear ICA takes the least time to compute the components, whereas the kernel method takes the highest time to compute the same number of components. In case of RFFICA, it is much faster than the kernel ICA method. For instance, when 10 components are considered our proposed method is $7X$ times faster than kernel ICA method, and this further increases when the number of components are increased. While compared with linear ICA, our method takes slightly more time, but it is negligible. This demonstrates that our method offers good classification perform with less computational burden.

TABLE IX: Comparison of classwise accuracy for 6 components of Pavia University dataset using LDA, GDA and RFFLDA

| # | Class | Accuracy(%) | | |
|---|---|---|---|---|
| | | LDA | GDA | RFFLDA |
| 1 | Asphalt | 67.97 | 71.89 | **81.09** |
| 2 | Meadows | 65.14 | 68.61 | **73.94** |
| 3 | Gravel | 70.99 | **78.54** | 76.85 |
| 4 | Trees | **95.82** | 93.56 | 93.78 |
| 5 | Painted metal sheets | 99.60 | 99.76 | **99.92** |
| 6 | Bare Soil | 76.71 | 78.80 | **87.99** |
| 7 | Bitumen | 91.38 | 92.68 | **93.54** |
| 8 | Self-Blocking Bricks | 73.73 | 74.61 | **75.63** |
| 9 | Shadows | 96.88 | 97.96 | **99.41** |
| | **Average Accuracy** | 82.02 | 84.04 | **86.90** |

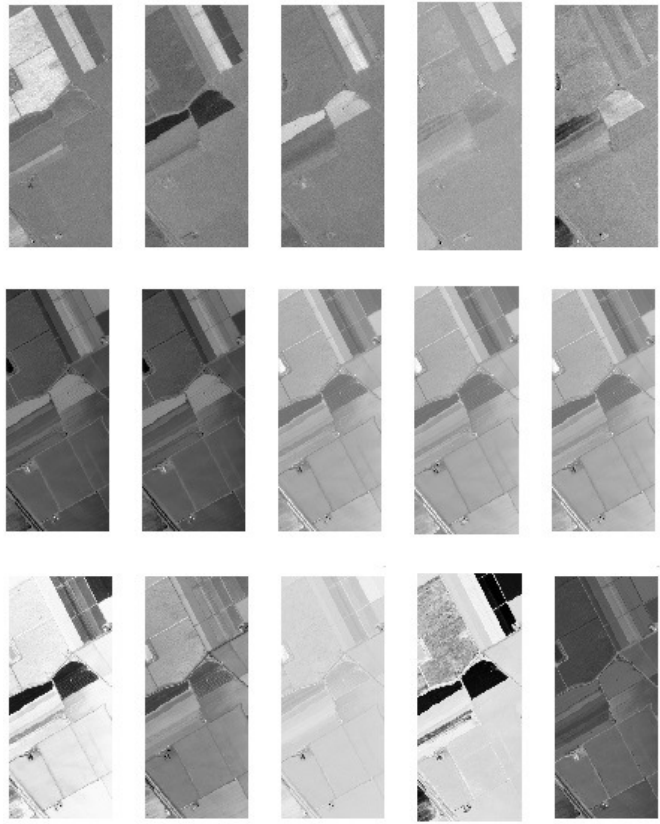

Fig. 6: The first five components of Salinas scene computed using: (a) LDA (first row from top), (b) GDA (middle row) and (c) RFFLDA (last row from top)

TABLE X: Comparison of Computational time (in Sec) required for 35 components of Salians scene using RFFICA against ICA and Kernel-ICA.

| Method | Computational Time Required (sec) Salinas Scene | | | | | | | |
|---|---|---|---|---|---|---|---|---|
| | 3 | 5 | 10 | 15 | 20 | 25 | 30 | 35 |
| ICA | 4.06 | 5.05 | 8.84 | 14.91 | 20.04 | 27.71 | 41.01 | 46.10 |
| KICA | 7.83 | 20.29 | 82.35 | 180.93 | 349.79 | 593.26 | 903.98 | 1688.80 |
| RFFICA | 7.77 | 6.76 | 13.60 | 19.44 | 28.57 | 35.99 | 47.23 | 53.61 |

### D. Impact of number of RFF features

In all the experiments so far, the number of RFF features is used with our proposed method is $2d$, where $d$ is the number of bands of HSI data. As it is known that when the number of RFF features is increased, it better approximates the kernel matrix [18]. Thus, in order to understand the impact of number of RFF features, we conducted experiments with varying number of RFF features with RFFLDA method and the results are shown in Fig. 7.

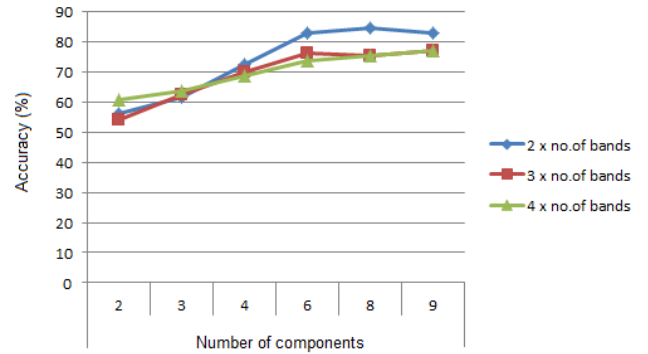

Fig. 7: Variation in classification accuracy (%) of RFFLDA against the different number of RFF features experimented on Pavia University dataset.

Fig. 7 confirms that the empirically chosen number of RFF feature expansion offers better classification performance. Increasing the RFF feature expansions doesn't lead to better performance, instead it decreases the classification performance in the higher order components.

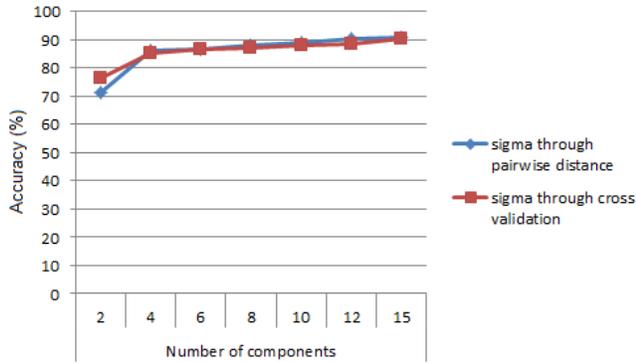

Fig. 8: Impact of bandwidth parameter of the Normal distribution used in the proposed RFFLDA for the Salians scene dataset

### E. Impact of bandwidth parameter ($\sigma$)

In our experiment, we empirically chose the value of the bandwidth parameter based on square of pairwise distance between the samples. In general, cross-validation approach is considered in literature to find the appropriate bandwidth parameter. In order to analyse the impact of bandwidth parameter on our proposed method, the $\sigma$ value was replaced by the optimal value obtained through five-fold cross validation method using all the band information. It can be observed from Fig. 8 that the performance of the empirical chosen value for $\sigma$ with our proposed method is close to the performance of the $\sigma$ selected through five-fold cross validation method.

## VI. Conclusion

In this paper, we proposed two randomized dimensionality reduction methods, namely randomized ICA and randomized LDA to reduce the dimensionality of the hyperspectral images. The proposed methods are developed based on the random Fourier features to represent the non-linearities present in the HSI data. The proposed method solved the shortcomings of the conventional kernel based dimensionality reduction methods (kernel ICA and kernel LDA), which cannot handle large amount of data and they rely on sub-optimal sampling strategy to compute the kernel matrix. Since few samples do not incorporate much information, the solution obtained through this will not be much precise. In contrast to this, our proposed method can scale to the large scale data, and can efficiently represent the non-linearities present in the data. Extensive experiments concluded that our proposed method extracts the informative features, as a result it outperforms the existing methods. Furthermore, it also reveals that our proposed methods were less computational burden than conventional kernel based method. Thus, our proposed method can be used as an replacement strategy for the kernel based methods to avoid the computational complexity and time.

The performance of our proposed method depends on the expressive power of RFF features, as the RFF coefficients are sampled in a data independent manner they might not have much expressive power. As a part of future work, randomized dimensionality reduction methods based on data dependent kernel approximation will be carried out.

## VII. Acknowledgement

The authors would like to thank Prof. P. Gamba for providing ROSIS hyperspectral images.